\documentclass[11pt]{article}
\usepackage{booktabs}
\usepackage{makecell}
\usepackage{booktabs}
\usepackage{siunitx}
\usepackage{array}
\usepackage{amsmath}
\usepackage{amssymb}
\usepackage{algorithm}
\usepackage{algpseudocode}
\usepackage{bbm}
\usepackage{graphicx} 
\usepackage{subcaption} 
\usepackage[style=apa]{biblatex}
\usepackage[colorlinks=true, linkcolor=blue, citecolor=blue, urlcolor=blue]{hyperref}

\usepackage[table]{xcolor}
\usepackage{slashbox}
\usepackage{setspace}
\usepackage[a4paper, margin=1in]{geometry}
\usepackage{multirow}
\usepackage{geometry}
\usepackage{amsthm}

\newtheorem{assumption}{Assumption}

\usepackage{newtxtext}
\usepackage{newtxmath}
\usepackage{bm}

\usepackage[margin=1in]{geometry}

\usepackage[font=small,labelfont=bf]{caption}
\usepackage{subcaption}

\usepackage[colorlinks=true,citecolor=blue,linkcolor=blue]{hyperref}

\usepackage{fancyhdr}
\title{Generative Adversarial Regression (GAR): Learning Conditional Risk Scenarios}
\author{
\large Saeed Asadi \hspace{3em} Jonathan Yu-Meng Li\\[1em]
\normalsize Telfer School of Management\\
University of Ottawa\\
\normalsize\texttt{sasad053@uottawa.ca, jonathan.li@telfer.uottawa.ca}
}

\begin{document}
\maketitle
\begin{abstract}
We propose Generative Adversarial Regression (GAR), a framework for learning conditional risk scenarios through generators aligned with downstream risk objectives. GAR builds on a regression characterization of conditional risk for elicitable functionals, including quantiles, expectiles, and jointly elicitable pairs. We extend this principle from point prediction to generative modeling by training generators whose policy-induced risk matches that of real data under the same context. To ensure robustness across all policies, GAR adopts a minimax formulation in which an adversarial policy identifies worst-case discrepancies in risk evaluation while the generator adapts to eliminate them. This structure preserves alignment with the risk functional across a broad class of policies rather than a fixed, pre-specified set. We illustrate GAR through a tail-risk instantiation based on jointly elicitable $(\mathrm{VaR}, \mathrm{ES})$ objectives. Experiments on S\&P 500 data show that GAR produces scenarios that better preserve downstream risk than unconditional, econometric, and direct predictive baselines while remaining stable under adversarially selected policies.

\end{abstract}

\section{Introduction}
Scenario generators are a standard tool in risk-sensitive decision pipelines. They produce plausible future trajectories, which are then propagated through downstream decision policies to generate outcomes on which a risk functional is evaluated. This workflow appears broadly across risk management applications, including finance, operations, and other settings where decisions must remain reliable under uncertain future trajectories.

A central requirement for scenario generators in practice is \emph{conditionality}: risk depends on the current state, and scenario generators are expected to adapt to covariates that summarize the relevant context. In financial risk management—our motivating application—this requirement underlies stress testing and regulatory emphasis on specific market regimes. The same principle applies more generally whenever risk varies systematically with observed conditions.

Beyond the challenges of generating high-dimensional objects such as trajectories and incorporating conditional information, scenario generators often suffer from \emph{risk misalignment}. In practice, generators are typically trained to make synthetic trajectories distributionally similar to historical ones under generic discrepancy criteria. Yet downstream risk is evaluated only after scenarios are mapped into outcomes by a policy, and distributional similarity at the trajectory level does not, in general, target the decision-relevant aspects of the conditional distribution that determine the resulting risk. This gap is particularly pronounced in high-dimensional settings and for risk functionals that emphasize rare-but-consequential events: discrepancies that appear negligible under standard distributional metrics can translate into materially incorrect risk assessments once a downstream policy extracts the relevant signal.

This paper proposes \emph{Generative Adversarial Regression (GAR)}, a risk-aligned framework for conditional scenario generation in high dimensions. GAR leverages a regression characterization of conditional risk, available for risk functionals satisfying the property of elicitability (\cite{Gneiting2011}). Rather than treating scenario generation and risk evaluation as separate stages, GAR defines the regression target directly as the policy-induced risk implied by the generator under the same context. In this way, the generator is trained to align its outputs with the downstream risk functional itself, rather than with an intermediate notion of distributional similarity.

A central bottleneck in this design is the choice of policies used to evaluate risk. Downstream policies are not fixed in practice—owing to re-tuning, changing constraints, or evolving objectives—and existing approaches often rely on guarantees tied to a finite, pre-specified set of policies. To overcome this limitation, GAR incorporates robustness to policy shift through a minimax formulation. The generator is trained against an adversarial policy that searches for worst-case discrepancies in risk evaluation, thereby ensuring risk-alignment beyond any fixed finite policy set.

\subsubsection*{Related work}

{\it Conditional generative models for time series.} Conditional simulation of time series is commonly approached through conditional generative models, most notably Conditional GANs and their variants. In finance, an early wave of work uses generative models to simulate market trajectories, learning the joint distribution of time-series paths from historical data (\cite{Tak19,Yoo19,Wie20}). For conditional generation, several papers adopt the Conditional-GAN framework (\cite{mirza2014conditionalgenerativeadversarialnets}) to condition on market covariates or state summaries (\cite{FuR19,Kos19,NiH,Vul24}). With the exception of approaches that incorporate domain-specific objectives (e.g., \cite{Vul24}), most of this literature trains generators using generic distributional discrepancy criteria such as Jensen--Shannon or Wasserstein-type divergences. While effective for matching overall distributional mass, such criteria are not tailored to downstream risk and may underweight rare but consequential events.

{\it Tail-aware modeling and evaluation.}
A complementary line of work targets heavy tails and extremes by blending generative modeling with tools from extreme value theory. For example, Pareto-GAN (\cite{Hus21}) leverages EVT to better match marginal tail behavior, but its performance can hinge on accurate tail-index estimation, which is statistically demanding in practice. More broadly, tail matching is not, by itself, a risk-aligned notion of fidelity: many risk functionals—such as expectiles or distortion-based criteria—depend on the full distribution, not merely its asymptotic tails. From a risk-management perspective, scenario generators should therefore be judged by whether they reproduce the distributional features that matter for the chosen risk criterion and downstream decisions, rather than by generic trajectory similarity or tail fit alone.

{\it Risk-aligned scoring for generators.}
Most closely related to our work, \cite{Con231} is an early effort to train market scenario generators directly against a risk functional. It fits an \emph{unconditional} GAN using strictly consistent scoring rules for the elicitable VaR/ES pair, with downstream tail-risk evaluations serving as the learning signal.  
The framework, however, remains unconditional and anchors calibration to an exogenously specified, fixed collection of benchmark policies, rendering performance sensitive to policy choice. More fundamentally for our setting, while scoring rules and elicitability are classical tools—widely used in the traditional literature on regression and forecast evaluation (\cite{Gneiting2011}), their use as a \emph{regression foundation for conditional generative modeling} has remained largely absent—in particular, to derive a regression characterization of \emph{conditional} risk and embed it directly in scenario generation. GAR fills this gap by exploiting elicitability to define a conditional regression target and coupling it with an \emph{adversarial} policy mechanism that identifies worst-case policies, yielding risk-aligned, context-aware scenario learning without reliance on a fixed benchmark set.




In the remainder of the paper, Sections~\ref{sec:gar} formalize conditional risk scenario generation and present GAR, which learns conditional scenarios via an elicitability-driven regression target and an adversarial policy mechanism to reduce dependence on a fixed benchmark set. Section~\ref{imp} details implementation and algorithms, Section~\ref{num} presents an empirical study on S\&P~500 data, and Section~\ref{con} concludes.

\section{Generative Adversarial Regression (GAR)}
\label{sec:gar}
We study conditional scenario generation for downstream risk-sensitive decision making. We observe a dataset
\[
\mathcal{D}=\{(c_i,y_i)\}_{i=1}^N
\]
consisting of independent draws from the joint distribution of random variables $(C,Y)$ defined on a probability space $(\Omega,\mathcal{F},\mathbb{P})$. Here, $C:\Omega \to \mathbb{R}^{d_{c}}$ denotes observable covariates (context), and $Y:\Omega \to \mathbb{R}^{M\times T}$ denotes a high-dimensional scenario. In financial applications, $Y$ represents a multivariate return or price trajectory over $T$ periods across $M$ assets, while $C$ summarizes prevailing market conditions. The formulation applies more broadly to any setting where high-dimensional trajectories are conditioned on observable context.

\subsection{Conditional Risk Scenario Generation} \label{sec:2.1}
The objective of conditional scenario generation is to learn a mechanism that, for each fixed $c$, produces samples from the conditional law $\mathcal{L}(Y\mid C=c)$. Formally, this corresponds to specifying a measurable mapping 
\[G: \Omega \times \mathbb{R}^{d_{c}} \to \mathbb{R}^{M\times T}\]
such that, for each fixed $c$, the random element $G(\cdot,c)$ has distribution $\mathcal{L}(Y\mid C=c)$.

In practice, we parameterize the generator $G$ through a latent-variable construction. Let $Z \sim F_Z$ be a random variable independent of $C$. A parametric conditional generator is then specified by 
\[G_\theta:\mathbb{R}^{d_z}\times \mathbb{R}^{d_{c}} \to \mathbb{R}^{M\times T},\]
and, given context $c$, synthetic scenarios are generated as
\[Y_{\theta} = G_\theta(Z,c).\]
Standard training objectives aim to match $\mathcal{L}(G_\theta(Z,c))$ to $\mathcal{L}(Y\mid C=c)$ under a chosen distributional discrepancy. However, in high-dimensional time-series settings, distribution matching is intrinsically difficult. More importantly, small discrepancies under generic metrics may translate into substantial downstream errors once decisions and risk evaluations are applied.

To formalize downstream relevance, let $\pi:\mathbb{R}^{M\times T}\to\mathcal{U}^{T}$ be a policy (decision rule)
that maps a scenario to an action sequence, and let $A:\mathbb{R}^{M\times T}\times\mathcal{U}^{T}\to\mathbb{R}$
be a deterministic aggregator.
Define a policy-induced measurable functional
\begin{equation}
\label{eq:policy}
    \Pi(Y) := A\big(Y,\pi(Y)\big), \qquad \Pi:\mathbb{R}^{M\times T}\to\mathbb{R}.
\end{equation}

For brevity, we refer to $\Pi$ as the policy-induced functional throughout, and denote $L_{\Pi} = \Pi(Y)$ as the real outcome. Given a conditional generator $G_\theta$, the same policy induces a synthetic outcome \[L_{\theta,\Pi} = \Pi(Y_\theta)=\Pi\left(G_\theta(Z,c)\right).\]


\paragraph{Elicitable risk functional} Let $\rho$ be a law-invariant risk functional of primary interest  (e.g., VaR/ES or an expectile). From a risk-management perspective, a conditional generator is meaningful only if it preserves policy-induced conditional risk, in the sense that 
\[
\rho(L_{\theta,\Pi}\mid C=c)\approx \rho(L_{\Pi}\mid C=c)\;\; {\rm for\; all\; relevant\;} c\; {\rm and\; admissible\;} \Pi.
\]
The central challenge is to achieve such risk alignment uniformly over contexts and across a broad class of policies. To proceed, we impose first a structural assumption on the risk functional.
\begin{assumption}[Elicitability; \cite{Gneiting2011}]
The risk functional $\rho$ is elicitable on a class of distributions $\mathcal{P}$. That is, there exists a scoring function
$S:\mathbb{R}^d\times\mathbb{R}\to\mathbb{R}$ such that for any $L\sim \mu\in\mathcal{P}$,
\begin{equation}
\rho(L)\in\arg\min_{a\in\mathbb{R}^d}\ \mathbb{E}\big[S(a,L)\big],
\label{eq:elicitable_def}
\end{equation}
and the score is strictly consistent if the minimizer is unique almost surely.
\end{assumption} 
Examples of elicitable risk functionals include quantiles, expectiles, and jointly elicitable pairs such as (VaR,ES).

\subsection{Elements of GAR} \label{sec:2.2}
We now outline GAR in three steps.

\paragraph{Step 1: Conditional risk as regression}
We first address the problem of learning conditional risk across different contexts $c$. Under the elicitability assumption, conditional risk admits a regression characterization. For a fixed policy-induced functional $\Pi$ and context $c$, applying elicitability conditionally yields
\begin{equation}
\rho(L_{\Pi}\mid C=c)\in\arg\min_{a\in\mathbb{R}^d}\ \mathbb{E}\big[S(a,L_{\Pi})\mid C=c\big].
\label{eq:elicitable_def_cond}
\end{equation}
Consider minimizing the unconditional expected score over measurable predictors $a(\cdot)$:
\begin{equation} 
\min_{a(\cdot)}\ \mathbb{E}\big[S(a(C),L_{\Pi})\big].
\label{eq:global_reg}
\end{equation}
By the law of iterated expectations, the objective decomposes pointwise in $c$. Strict consistency therefore implies that any minimizer satisfies
$$a(c) = \rho(L_{\Pi}\mid C=c)\;{\rm for\;almost\;every\;}c.$$
Hence, learning the conditional risk function $c \mapsto \rho(L_{\Pi}\mid C=c)$ reduces to solving a regression problem \eqref{eq:global_reg} under the strictly consistent score.

\paragraph{Step 2: Generative regression}
Our objective, however, is not merely to estimate the conditional risk $a(c)$, but to train a \emph{conditional generator} $G_\theta$ whose induced outcomes reproduce the correct downstream risk. Fix a policy-induced functional $\Pi$. For each context $c$, the generator $G_\theta$ produces a synthetic 
scenario $G_\theta(Z,c)$, which induces the synthetic outcome $\Pi\left(G_\theta(Z,c)\right)$. The generator therefore determines the conditional risk
\begin{equation}
a_{\theta,\Pi}(c)
:=\rho\big(\Pi(G_\theta(Z,C)) \mid C=c\big).
\label{eq:a_theta_pi}
\end{equation}
Embedding this generator-implied predictor into the regression formulation \eqref{eq:global_reg}  yields the generative regression objective
\begin{equation}
\min_{\theta}\
\mathbb{E}\Big[
S\big(a_{\theta,\Pi}(C),\ L_{\Pi}\big)
\Big]
=
\min_{\theta}\
\mathbb{E}\Big[
S\Big(
\rho\big(\Pi(G_\theta(Z,C))\mid C\big),
\ \Pi(Y)
\Big)
\Big].
\label{eq:gen_reg_obj}
\end{equation}
This objective trains the generator $G_\theta$ to align the generator-implied conditional risk $\rho(L_{\theta,\Pi}\mid C)$ with the true conditional risk $\rho(L_\Pi\mid C)$ under the strictly consistent score.

A remaining challenge is to ensure risk alignment not only for a fixed $\Pi$ but across a class of admissible policies. A natural extension averages the objective over a finite benchmark set $\{\Pi_k\}_{k=1}^K$:
\begin{equation}
\min_{\theta}\ \frac{1}{K}\sum_{k=1}^K
\mathbb{E}\Big[
S\Big(
\rho\big(\Pi_k(G_\theta(Z,C))\mid C\big),\ \Pi_k(Y)
\Big)
\Big].
\label{eq:finite_policy_obj}
\end{equation}
While this enforces risk alignment for the chosen benchmark strategies, performance remains guaranteed only for that fixed collection and may not generalize beyond it.


\paragraph{Step 3: Adversarial policy robustification (GAR)}
To address this limitation, GAR replaces the finite benchmark set with a rich admissible policy class and trains the generator against the worst-case policy in that class.

Let $\{\Pi_\phi\}_{\phi\in\Phi}$ denote a parameterized family of admissible policy-induced functionals. GAR solves the minimax problem
\begin{equation}
\min_{\theta}\ \max_{\phi\in\Phi}\
\mathbb{E}\Big[
S\Big(
\rho\big(\Pi_\phi(G_\theta(Z,C))\mid C\big),\ \Pi_\phi(Y)
\Big)
\Big].
\label{eq:gar_minimax}
\end{equation}
The inner maximization identifies policies under which the discrepancy between real and synthetic conditional risk is largest under the score $S$, while the outer minimization adjusts the generator to eliminate these discrepancies. Consequently, the learned generator preserves policy-induced conditional risk uniformly over contexts and across the admissible policy class, rather than being calibrated to a fixed, potentially brittle set of policies.


\section{End-to-End Training and Optimization of GAR}
\label{imp}
\begin{figure}[t]
  \centering
  \begin{subfigure}{\linewidth}
    \centering
    \includegraphics[width=\linewidth]{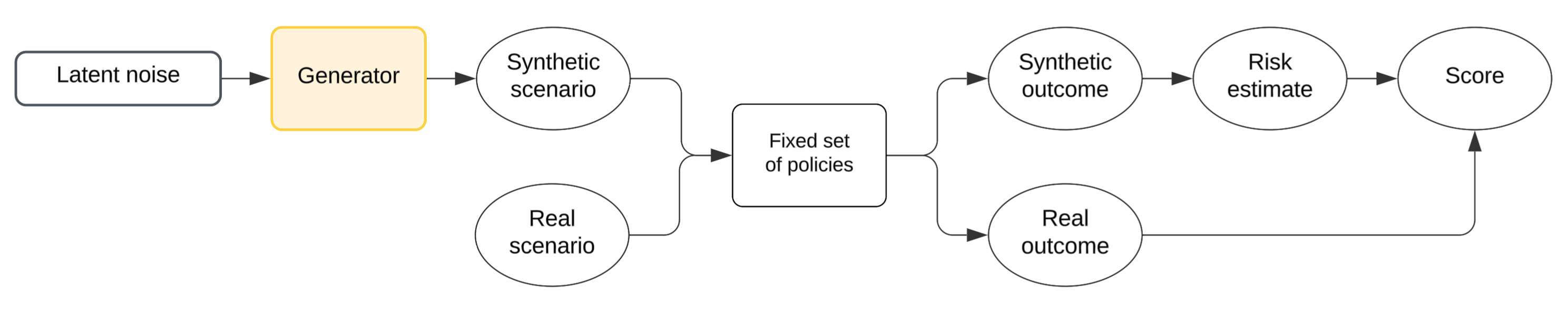}
    \caption{Baseline: unconditional + fixed trading strategies}
    \label{fig:pipeline:baseline}
  \end{subfigure}

  \vspace{0.8em}

  \begin{subfigure}{\linewidth}
    \centering
    \includegraphics[width=\linewidth]{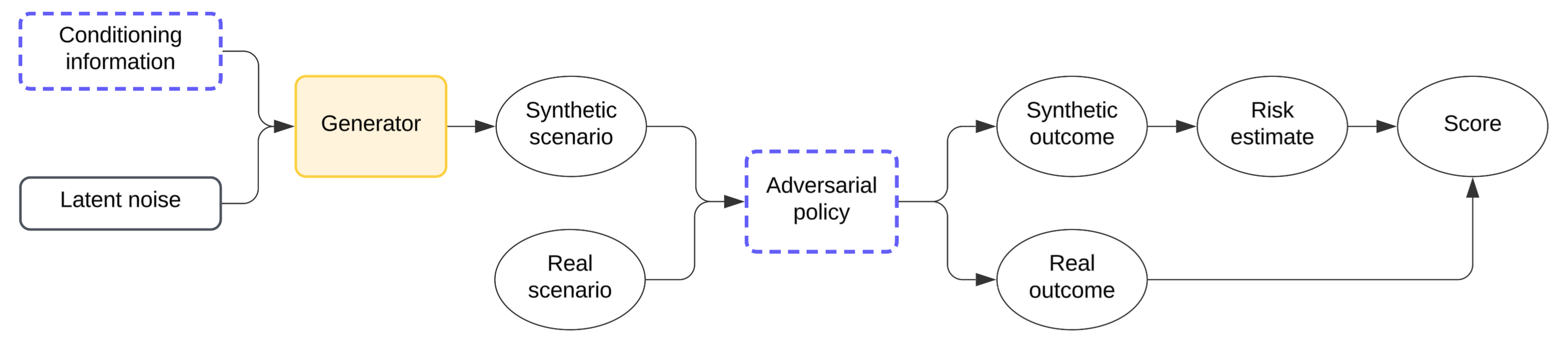}
    \caption{GAR}
    \label{fig:pipeline:robust}
   \end{subfigure}

  \caption{
  Comparison of training pipelines.
  (\subref{fig:pipeline:baseline}) Baseline framework: an unconditional generator produces scenarios that are evaluated under a fixed set of policies to obtain outcomes.
  (\subref{fig:pipeline:robust}) Proposed framework: the generator conditions on context and is trained in a min--max game against an adversarial policy, encouraging risk estimates that are robust to worst-case policies.
  }
  \label{fig:pipeline:comparison}
\end{figure}

In this section, we operationalize the three core elements of GAR and describe the end-to-end procedure used to solve the minimax objective in \eqref{eq:gar_minimax}. Specifically, we detail (i) the choice of elicitable risk targets and strictly consistent scoring rules, (ii) the estimation of generator-implied conditional risk, and (iii) the adversarial policy class together with alternating stochastic updates for the GAR minimax objective. To highlight the distinction between GAR and existing unconditional approaches calibrated to a fixed set of benchmark policies, Figure~\ref{fig:pipeline:comparison} contrasts the corresponding training pipelines.

\subsection{Risk Targets and Strictly Consistent Scores}
In this subsection, we specify the scoring functions used to estimate the conditional risk $\rho(L \mid C=c)$ in~\eqref{eq:elicitable_def_cond}.

We focus on three risk functionals: Value at Risk (VaR), Expected Shortfall (ES), and the
expectile. For a scalar random variable $L$ with cumulative distribution function $F_L$,
their definitions are
\[
VaR_\alpha(L) \;:=\; \inf\big\{\,\ell\in\mathbb{R}: F_L(\ell)\ge \alpha\,\big\},
\qquad
ES_\alpha(L) \;:=\; \frac{1}{\alpha}\int_{0}^{\alpha}VaR_\beta(L)\, d\beta.
\]
The $\tau$-expectile is defined as the unique solution $m\in\mathbb{R}$ to
\[
\tau\,\mathbb{E}\big[(L-m)^+\big] \;=\; (1-\tau)\,\mathbb{E}\big[(L-m)^-\big],
\]
where $(x)^+ := \max\{x,0\}$ and $(x)^- := \max\{-x,0\}$.

Let $\ell$ denote a realization of $L$. Strictly consistent scores for scalar elicitable functionals include the quantile (VaR) score
\begin{equation}
S_\alpha(a,\ell)
\;=\;
\bigl|\alpha-\mathbbm{1}_{\{\ell\le a\}}\bigr|\,|\ell-a|,
\label{eq:var_score}
\end{equation}
and the expectile score
\begin{equation}
S_\tau(a,\ell)
\;=\;
\bigl|\tau-\mathbbm{1}_{\{\ell\le a\}}\bigr|\,(\ell-a)^2.
\label{eq:expectile_score}
\end{equation}
However, $ES_\alpha$ is not elicitable on its own, whereas the pair
$(VaR_\alpha,ES_\alpha)$ is jointly elicitable and admits strictly consistent scoring functions. A broad class of such scores is given by~\cite{fissler2015expectedshortfalljointlyelicitable}:
\begin{align}
S_\alpha\big(a,\ell\big)
&=
\left(\mathbbm{1}_{\{\ell \le v\}}-\alpha\right)\big(H_1(v)-H_1(\ell)\big)
+\frac{1}{\alpha} H_2'(e)\,\mathbbm{1}_{\{\ell \le v\}}(v-\ell) \nonumber\\
&\quad + H_2'(e)(e-v)-H_2(e),
\label{eq:score_joint}
\end{align}
where $a:=(v,e)$, $H_1$ and $H_2$ are functions with $H_1$ increasing, and $H_2$ differentiable, strictly increasing,
and strictly convex.

In our experiments, we follow~\cite{fissler2015expectedshortfalljointlyelicitable} and set
$H_1(v)=v$ and $H_2(e)=s\exp(e/s)$ with scale $s>0$. See~\cite{acerbi2014} for an
alternative specification of $(H_1,H_2)$.

\paragraph{Remark}
In~\eqref{eq:score_joint}, the indicator $\mathbbm{1}_{\{\ell \le v\}}$ is non-differentiable, so we replace it with the smooth surrogate
\begin{equation}
\mathbbm{1}_{\{\ell \le v\}} \approx \sigma_k(v-\ell)
:=\frac{1}{1+\exp(-k(v-\ell))},
\label{eq:sigmoid_surrogate}
\end{equation}
with sharpness parameter $k>0$, enabling stable backpropagation through the score.

\subsection{Estimation of Generator-Implied Conditional Risk}
For a fixed context $c$ and policy-induced functional $\Pi_\phi$, we approximate the generator-implied conditional distribution of the synthetic outcome $\Pi_\phi(G_\theta(Z,c))$ via Monte Carlo sampling. Specifically, for a sample size $N_{MC}$, draw i.i.d.\ latent samples $z_{1},\dots,z_{N_{MC}} \sim F_Z$ and form synthetic scenarios $\hat y_{s} = G_\theta(z_{s},c)$ and outcomes $\hat l_{s}=\Pi_\phi(\hat y_{s})$, $s=1,\dots,N_{MC}$.

We then estimate the generator-implied conditional risk
\[
a_{\theta,\Pi_\phi}(c) := \rho\!\left(\Pi_\phi(G_\theta(Z,C)) \mid C=c\right)\in\mathbb{R}^d
\]
by a plug-in estimator computed from the empirical distribution of
$\{\hat l_{s}\}_{s=1}^{N_{MC}}$, yielding $\hat a_{\theta,\Pi_\phi}(c)=\rho(\{\hat l_{s}\}_{s=1}^{N_{MC}})$ (with a slight abuse of notation, we use $\rho$ both for the risk functional on distributions and for its plug-in version acting on sample).
For $(\mathrm{VaR}_\alpha,\mathrm{ES}_\alpha)$ this corresponds to the empirical quantile and tail average,
respectively. The estimate $\hat a_{\theta,\Pi_\phi}(c)$ is the input to the scoring function in~\eqref{eq:gar_minimax} i.e. 
\begin{equation}
\min_{\theta}\ \max_{\phi\in\Phi}\
\mathbb{E}\Big[
S\Big(
\hat a_{\theta,\Pi_\phi}(C),\ \Pi_\phi(Y)
\Big)
\Big].    
\end{equation}

\subsection{Policy Class and Alternating Min--Max Optimization}
\paragraph{Policy parameterization.}
Recall~\eqref{eq:policy}: for any policy $\pi$, the scalar outcome is $\Pi(Y)=A\big(Y,\pi(Y)\big)$. In our adversarial instantiation, given a scenario $y\in\mathbb{R}^{M\times T}$, the policy $\pi_\phi$ outputs an action sequence
\begin{equation}
\label{eq:policy_seq}
    \pi_\phi(y) =
\big[w_\phi(y_{1:1}),\; w_\phi(y_{1:2}),\;\dots,\; w_\phi(y_{1:T-1})\big],
\qquad
w_\phi(\cdot)\in\mathbb{R}^M,
\end{equation}
where $w_\phi$ is \emph{causal} (non-anticipative), i.e., at time $t$ it depends only on the history $y_{1:t}$. Such a mapping can be parameterized by a standard temporal model (e.g., a recurrent model); details of our implementation are given in Appendix~\ref{app-B}.

Applying the deterministic aggregator $A$, we obtain the policy-induced functional
\begin{equation}
\Pi_\phi(y)
=
A\big(y,\pi_\phi(y)\big)
:=
\sum_{t=1}^{T-1}\sum_{j=1}^M \big(w_\phi(y_{1:t})\big)_j\,\Delta y_{j,t},
\qquad
\Delta y_{j,t}:=y_{j,t+1}-y_{j,t},
\label{eq:policy_outcome}
\end{equation}
where $\big(w_\phi(y_{1:t})\big)_j$ denotes the $j$-th component of $w_\phi(y_{1:t})\in\mathbb{R}^M$.

To avoid degenerate adversarial policies that exploit unbounded leverage, we constrain the policy’s outputs to have a fixed total gross exposure $\kappa>0$,
\begin{equation}
  \big\|w_\phi(y_{1:t})\big\|_1 = \kappa, \qquad t = 1,\dots,T-1,
  \label{eq:exposure_constraint}
\end{equation}
In practice, we obtain $w_\phi(y_{1:t})$ by taking the unconstrained output $h_t$ of the policy model and applying the normalization
\[
w_\phi(y_{1:t}) = \kappa\, \frac{h_t}{\|h_t\|_1},
\]
which enforces \eqref{eq:exposure_constraint}.

\paragraph{Optimization (alternating stochastic min--max).}
We optimize \eqref{eq:gar_minimax} via alternating stochastic gradient steps. For a minibatch
$\{(c_i,y_i)\}_{i=1}^B$, we (i) update the policy parameters $\phi$ to increase the score discrepancy,
\[
\phi \leftarrow \phi + \eta_\phi \nabla_\phi
\frac{1}{B}\sum_{i=1}^B
S\!\left(\hat a_{\theta,\Pi_\phi}(c_i),\ \Pi_\phi(y_i)\right),
\]
and then (ii) update the generator parameters $\theta$ to decrease it,
\[
\theta \leftarrow \theta - \eta_\theta \nabla_\theta
\frac{1}{B}\sum_{i=1}^B
S\!\left(\hat a_{\theta,\Pi_\phi}(c_i),\ \Pi_\phi(y_i)\right),
\]
where $\hat a_{\theta,\Pi_\phi}(c)$ is the Monte Carlo estimate of the generator-implied conditional risk. The min–max optimization procedure is detailed in Algorithm~\ref{alg:scenario}.


\begin{algorithm}[H]
\caption{Alternating stochastic min--max training (GAR)}
\label{alg:scenario}
\begin{algorithmic}[1]
\Require Training set $\mathcal{D}_{\mathrm{trn}}$, learning rates $\eta_\theta,\eta_\phi$, Monte Carlo sample size $N_{MC}$
\State Initialize $\theta\leftarrow\theta_0$, $\phi\leftarrow\phi_0$
\For{epoch $n=1$ to $N$}
  \For{minibatch $\mathcal{B}=\{(c_i,y_i)\}\subset\mathcal{D}_{\mathrm{trn}}$}

    \Statex \hspace{-1.2em}\textbf{Adversary update (ascent; freeze $\theta$)}
    \State \textbf{Freeze} $\theta$
    \For{each $(c_i,y_i)\in\mathcal{B}$}
      \For{$s=1$ to $N_{MC}$}
            \State Draw $z_{s} \sim F_Z$
            \State Generate synthetic scenarios $\hat y_{i,s}\leftarrow G_\theta(z_{s},c_i)$
            \State Compute synthetic outcomes $\hat l_{i,s}\leftarrow \Pi_\phi(\hat y_{i,s})$
        \EndFor
      \State Estimate risk $\hat a_i \leftarrow \rho\!\left(\{\hat l_{i,s}\}_{s=1}^{N_{MC}}\right)$  
      \State Compute real outcomes $l_i\leftarrow \Pi_\phi(y_i)$
      \State Compute score $\ell_i \leftarrow S(\hat a_i, l_i)$
    \EndFor
    \State $\hat{\mathcal{L}}(\phi)\leftarrow \frac{1}{|\mathcal{B}|}\sum_{(c_i,y_i)\in\mathcal{B}}\ell_i$
    \State $\phi \leftarrow \phi + \eta_\phi \nabla_\phi \hat{\mathcal{L}}(\phi)$

    \Statex \hspace{-1.2em}\textbf{Generator update (descent; freeze $\phi$)}
    \State \textbf{Freeze} $\phi$
    \For{each $(c_i,y_i)\in\mathcal{B}$}
        \For{$s=1$ to $N_{MC}$}
            \State Draw $z_{s} \sim F_Z$
            \State Generate synthetic scenarios $\hat y_{i,s}\leftarrow G_\theta(z_{s},c_i)$
            \State Compute synthetic outcomes $\hat l_{i,s}\leftarrow \Pi_\phi(\hat y_{i,s})$
        \EndFor
      \State Estimate risk $\hat a_i \leftarrow \rho\!\left(\{\hat l_{i,s}\}_{s=1}^{N_{MC}}\right)$  
      \State Compute real outcomes $l_i\leftarrow \Pi_\phi(y_i)$
      \State Compute score $\ell_i \leftarrow S(\hat a_i, l_i)$
    \EndFor
    \State $\hat{\mathcal{L}}(\theta)\leftarrow \frac{1}{|\mathcal{B}|}\sum_{(c_i,y_i)\in\mathcal{B}}\ell_i$
    \State $\theta \leftarrow \theta - \eta_\theta \nabla_\theta \hat{\mathcal{L}}(\theta)$

  \EndFor
\EndFor
\end{algorithmic}
\end{algorithm}

\begin{figure}[h]
    \centering
    \begin{subfigure}{0.48\linewidth}
        \centering
        \includegraphics[width=\linewidth]{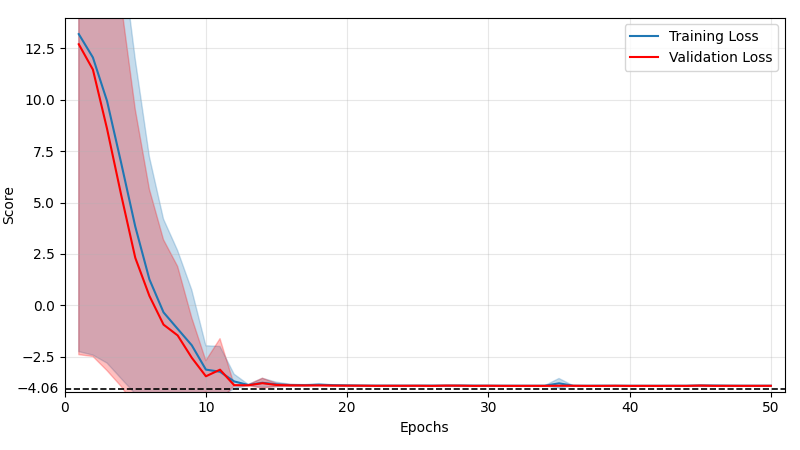}
        \caption{Encoder–Linear}
        \label{fig:loss-enc}
    \end{subfigure}\hfill
    \begin{subfigure}{0.48\linewidth}
        \centering
        \includegraphics[width=\linewidth]{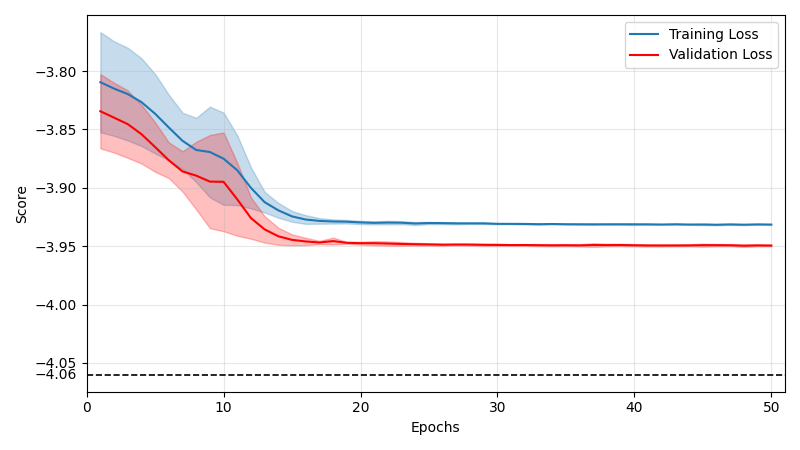}
        \caption{Encoder–LSTM}
        \label{fig:loss-lstm}
    \end{subfigure}
    \caption{Training/validation score curves for two generator architectures. Each experiment is repeated five times with different random seeds. The dotted line indicates the lowest attainable in-sample score. Scores are visualized with mean (solid lines) and standard deviation (shaded areas).}
    \label{fig:loss-both}
\end{figure}

\section{Numerical Experiments} \label{num}
We evaluate three conditional generator architectures $G_\theta$ (Encoder--LSTM, Encoder--Linear, and Simple--Linear; Appendix~\ref{app-B}) in two complementary settings. First, to quantify the impact of conditionality, we fix two classes of policies and train the conditional generators under the fixed-policy objective \eqref{eq:finite_policy_obj}, then compare against the baseline models; results are summarized in Table~\ref{tab:results}. Second, to study robustness to policy shift, we activate the min--max framework by treating the policy stage as an adversary and train the conditional generators via \eqref{eq:gar_minimax}, comparing them with their fixed-policy counterparts trained under \eqref{eq:finite_policy_obj}; results are reported in Table~\ref{tab:results_robust}.

\begin{table}[h]
\centering
\begin{tabular}
{l
                S[table-format=-1.3]
                S[table-format=-1.3]
                S[table-format=-1.3]
                S[table-format=2.1]}
\toprule
\multirow{2}{*}{Model}
  & \multicolumn{3}{c}{VaR–ES score $S_{\alpha}$ (↓)} & {VaR viol.\ rate (\%)} \\
\cmidrule(lr){2-4}
  & {Train} & {Validation} & {Test} & {(at $\alpha=5\%$)} \\
\midrule
Unconditional    & -3.747 & -3.806 & -3.765 & 1.2 \\
DCC--GARCH       & -3.545 & -3.754 & -3.774 & 14.8 \\
Direct           & -3.669 & -3.744 & -3.760 & 1.5 \\
\midrule
Encoder--LSTM    & \textbf{-3.932} & \textbf{-3.950} & \textbf{-3.929} & 6.6 \\
Encoder--Linear  & -3.911 & -3.922 & -3.920 & 2.5 \\
Simple--Linear   & -3.887 & -3.922 & -3.922 & 2.1 \\
\bottomrule
\end{tabular}
\caption{Comparison of conditional generators and baselines on the joint VaR--ES score $S_{\alpha}$ (Boldface indicates the best value in each column) and the VaR violation rate at $\alpha=5\%$ (closer to $5\%$ indicates better calibration). The violation rate is the fraction of periods in which the realized PnL falls below the model’s $\mathrm{VaR}_{0.05}$.}
\label{tab:results}
\end{table}

In our financial risk-management application, the policy specializes to a trading strategy applied to a price/return trajectory, and the resulting outcome is the portfolio profit-and-loss (PnL). We use daily log-returns of the following S\&P\,500 stocks: AAPL, INTC, T, F, BAC, NEE, MU, AMD, PFE, from 1984-06-01 to 2025-08-20, and split the dataset into training, validation, and test sets with a 80--10--10 ratio. Each sample is formed by taking the preceding 5 daily returns as conditioning information and the subsequent 10-day return path as the realized scenario. We evaluate all models using the strictly consistent joint scoring function \(S_{\alpha}(v,e,x)\) in \eqref{eq:score_joint} for (VaR, ES), and additionally report the VaR violation rate.

\begin{table}[h]
  \centering
  \begin{tabular}{>{\centering\arraybackslash}m{2.2cm} >{\centering\arraybackslash}m{2.2cm} 
                  >{\centering\arraybackslash}m{1.8cm} >{\centering\arraybackslash}m{1.8cm} >{\centering\arraybackslash}m{1.8cm} 
                  >{\centering\arraybackslash}m{1.8cm}}
  \toprule
  \multicolumn{2}{c}{} & \multicolumn{2}{c|}{Training Phase} & \multicolumn{2}{c}{Evaluation Phase} \\
  \cmidrule(r){3-4} \cmidrule(r){5-6}
  \multicolumn{2}{c}{} & Train & Validation & Benchmark & Worst-Case \\
  \midrule
  \multirow{2}{*}{Encoder-LSTM} 
    & Adversarial  & -3.864 & -3.885 & -3.921 & \textbf{-3.888} \\
    & Fixed        & -3.932 & -3.950 & \textbf{-3.929} & -3.558 \\
  \midrule
  \multirow{2}{*}{Encoder-Linear} 
    & Adversarial  & -3.762 & -3.806 & \textbf{-3.921} & \textbf{-3.781} \\
    & Fixed        & -3.911 & -3.922 & -3.920 & -1.349 \\
  \midrule
  \multirow{2}{*}{Simple-Linear} 
    & Adversarial  & -3.799 & -3.858 & -3.855 & \textbf{-3.761} \\
    & Fixed        & -3.887 & \-3.922 & \textbf{-3.922} &  1.233 \\
  \bottomrule
  \end{tabular}
  \caption{Performance of conditional generators under \emph{Fixed} and \emph{Adversarial} training using the joint VaR--ES score $S_{\alpha}(\cdot)$ (lower is better). Columns under \emph{Training Phase} report train/validation scores obtained during each model’s own training regime (fixed-policy objective \eqref{eq:finite_policy_obj} for \emph{Fixed}, min--max objective \eqref{eq:gar_minimax} for \emph{Adversarial}). Columns under \emph{Evaluation Phase} report test scores under benchmark (mean-reversion and trend-following) and worst-case strategies. Boldface in the Evaluation Phase highlights the better value (lower score) between \emph{Fixed} and \emph{Adversarial} for each architecture.}
  \label{tab:results_robust}
\end{table}

\paragraph{Conditionality impact}
To quantify the value of conditioning information, we fix two classes of trading strategies---mean reversion and trend following---and compare the conditional generators with three baseline models: an unconditional generator, a Dynamic Conditional Correlation GARCH (DCC--GARCH) model, and a direct (scenario-free) linear model.

The unconditional generator is trained using the framework demonstrated in Figure \ref{fig:pipeline:baseline}. Mathematically,
\begin{align}
\mathbf{G}^* \in \arg \min_{G_{\theta}} \frac{1}{K} \sum_{k=1}^{K}
\mathbb{E}
\big[ S_\alpha\!\big(v(\Pi_k(G_{\theta}(Z))),\, e(\Pi_k(G_{\theta}(Z))),\, \Pi_k(Y) \big) \big].
\end{align}

This baseline model allows us to evaluate the contribution of conditioning information to scenario generation and risk estimation. Our second baseline is DCC--GARCH model (\cite{Engle01072002}), a classical econometric approach for time-varying conditional covariance modeling. We fit univariate GARCH(1,1) models to each asset to capture idiosyncratic volatility dynamics, and then estimate a dynamic correlation structure to model evolving cross-asset dependence. We adopt the (1,1) specification because, among the alternative GARCH$(p,q)$ configurations we considered, it achieved the lowest Bayesian Information Criterion (BIC) while remaining parsimonious and standard in empirical risk-management practice. The resulting parametric model is used to simulate multivariate return scenarios.

To disentangle the benefit of conditional \emph{scenario generation} from simply using conditioning information, we also consider a baseline that directly maps the same conditioning information to risk estimates. We call this model a direct linear model. Specifically, we fit two linear regressions to predict VaR and ES at quantile level $\alpha$. Let $c \in \mathbb{R}^{M \times T}$ denote the conditioning information across $M$ assets and $T$ time steps, and let ${c}^{\mathrm{vec}} \in \mathbb{R}^D$ be its vectorization with $D = M \times T$. For $K$ strategies, the model estimates
\begin{align}
\widehat{\text{VaR}}_\alpha({c}) = {c}^{\mathrm{vec} \, \top} \mathbf{W}_{\text{VaR}} + \mathbf{b}_{\text{VaR}}, \quad
\widehat{\text{ES}}_\alpha({c}) = {c}^{\mathrm{vec} \, \top} \mathbf{W}_{\text{ES}} + \mathbf{b}_{\text{ES}},
\end{align}
where $\mathbf{W}_{\text{VaR}}, \mathbf{W}_{\text{ES}} \in \mathbb{R}^{D \times K}$ and $\mathbf{b}_{\text{VaR}}, \mathbf{b}_{\text{ES}} \in \mathbb{R}^K$. We use the same conditioning information as for the target conditional generators.

Figure~\ref{fig:loss-both} shows the training and validation loss curves for the Encoder--LSTM and Encoder--Linear architectures. The dotted horizontal line represents an oracle benchmark, computed under the idealization that the estimated VaR and ES coincide with the realized outcomes ($v_i = e_i = l_i$ for all $i$). Under this assumption, the scoring function $S_\alpha((v_i,e_i),l_i)$ reduces to $-2\exp(l_i/2)$, yielding the empirical mean
$
\mathcal{L}_{\min}
=
-\frac{2}{N}
\sum_{i=1}^{N}
\exp(l_i/2)
$
over the training sample. This quantity represents the lowest attainable in-sample score under perfect forecasts and serves as a benchmark for assessing convergence quality.

Table~\ref{tab:results} shows that all three conditional generators (Encoder--LSTM, Encoder--Linear, Simple--Linear) outperform the baselines on the joint VaR--ES score across the training, validation, and test splits. The consistent gap between the conditional generators and the unconditional generator highlights the value of market context for extreme-scenario generation. Encoder--LSTM achieves the best overall score, suggesting that explicitly modeling temporal structure improves risk-sensitive scenario generation.

Among the baselines, the unconditional generator performs best on the training and validation splits, while DCC--GARCH is slightly stronger on the test split. The direct linear model, which uses the same conditioning information but does not generate scenarios, is consistently weaker than the unconditional generator and is also outperformed by DCC--GARCH on validation and test, highlighting the benefit of distributional (scenario-based) modeling for conditional risk estimation.

For VaR calibration at $\alpha=5\%$, a well-calibrated $\mathrm{VaR}_{0.05}$ should be violated about $5\%$ of the time. Violation rates below $5\%$ indicate conservative VaR estimates (tail-risk overestimation), whereas rates above $5\%$ indicate aggressive VaR estimates, i.e., tail-risk underestimation. In Table~\ref{tab:results}, Encoder--LSTM records a $6.6\%$ violation rate, which is slightly above target but closest to $5\%$. DCC--GARCH has a much higher rate ($14.8\%$), indicating substantial tail-risk underestimation. The remaining models are conservative: Encoder--Linear ($2.5\%$) and Simple--Linear ($2.1\%$) are closest among them, followed by Direct ($1.5\%$) and Unconditional ($1.2\%$).
\begin{figure}[h]
    \centering
    \includegraphics[width=\linewidth]{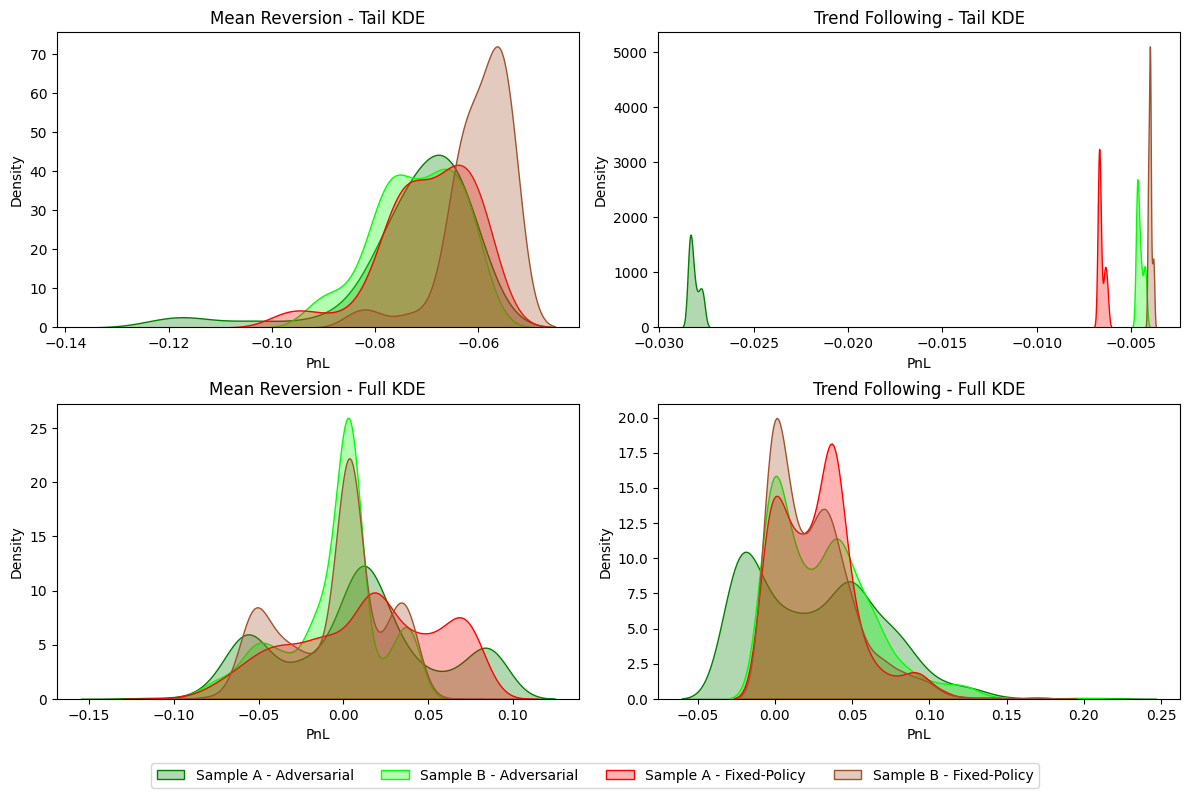}
    \caption{Sensitivity to conditioning information at PnL-level. Kernel density estimates of PnL for two strategy classes (mean reversion, trend following) under two market conditioning sample A and B, for fixed-policy and adversarial-policy training. Bottom panels: full PnL distribution; top panels: left tail (5\% quantile region).}
    \label{fig:sensitivity-pnl}
\end{figure}
\begin{figure}[h]
    \centering
    \includegraphics[width=0.8\linewidth]{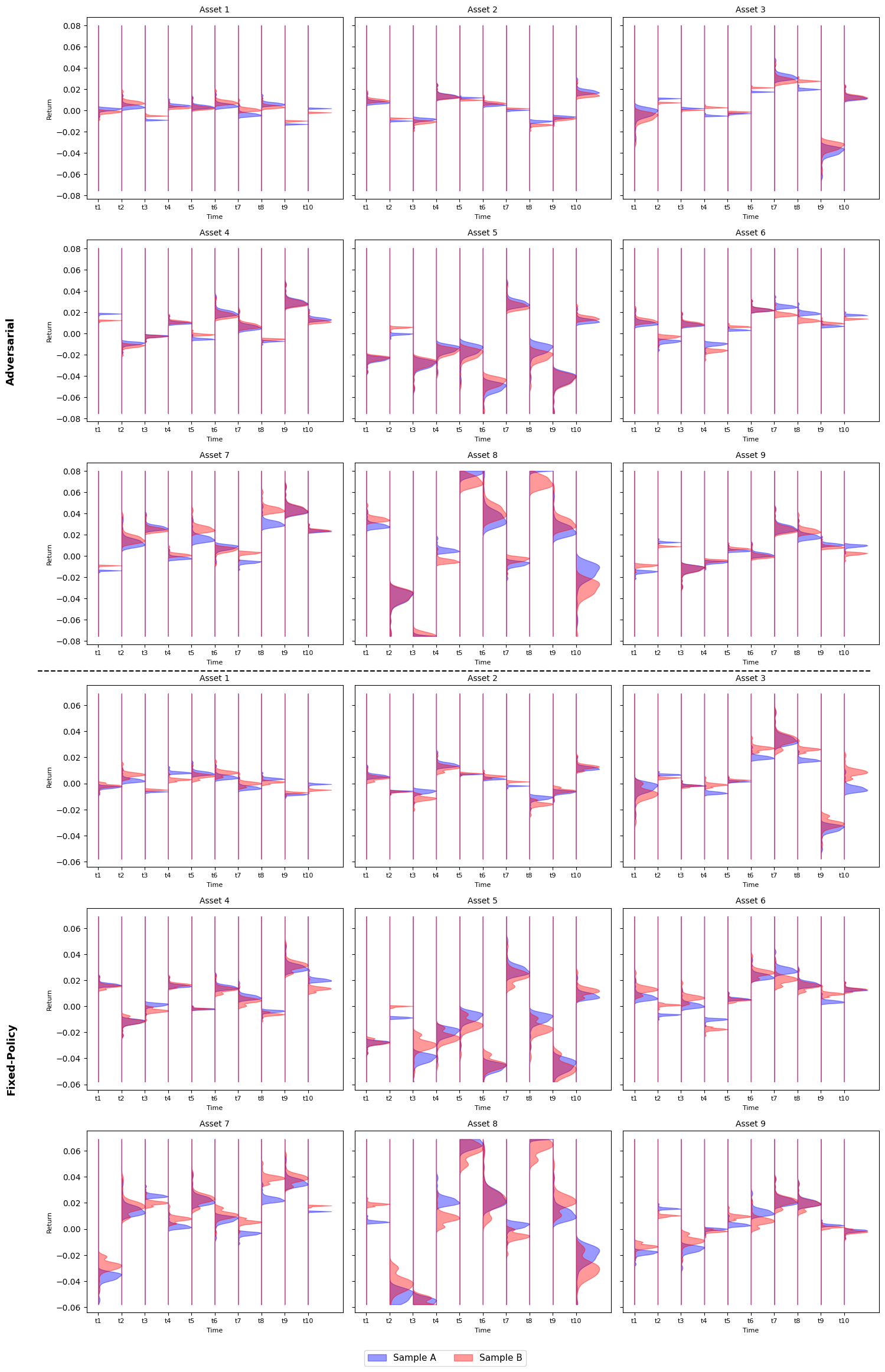}
    \caption{Sensitivity to conditioning information at Trajectory-level. For each asset and time step, densities of extreme scenarios contributing to the left tail of the PnL under two conditioning samples A and B, for adversarial-policy (top rows) and fixed-policy (bottom rows) training.}
    \label{fig:sensitivity-traj}
\end{figure}
\paragraph{Adversarial policy training}
Table~\ref{tab:results_robust} compares three conditional generator architectures trained under (i) the fixed-policy objective \eqref{eq:finite_policy_obj} and (ii) the adversarial (min--max) objective \eqref{eq:gar_minimax}, and evaluates each model on benchmark strategies (mean reversion and trend following) and worst-case strategies. On benchmark strategies, the two training regimes achieve similar performance, and adversarial training is not consistently better across architectures. On worst-case strategies, however, adversarially trained models consistently outperform their fixed-policy counterparts in all three architectures, with larger gains for the simpler architectures and a smaller gap for Encoder--LSTM. Overall, these results indicate that the main benefit of adversarial training is robustness to policy shift rather than universal improvement on benchmark strategies, while the stronger stability of Encoder--LSTM suggests better inherent generalization from a more expressive temporal architecture.

To illustrate how conditioning and training regime affect the generator and downstream risk, Figures \ref{fig:sensitivity-pnl} and \ref{fig:sensitivity-traj} examine sensitivity at two levels. Figure \ref{fig:sensitivity-pnl} shows the PnL level, with kernel density estimates of PnL under two strategy classes (mean reversion, trend following) and two conditioning samples, zooming on the full distribution and the left tail to show how conditioning and training regime alter tail behavior. Figure \ref{fig:sensitivity-traj} moves to the trajectory-level: for each asset and time step, we overlay densities of extreme scenarios that drive the left tail of the PnL distribution under two conditioning states, comparing fixed-policy and adversarial-policy training.

\section{Conclusion} \label{con}
We proposed Generative Adversarial Regression (GAR), a framework for conditional scenario generation that aligns generative objectives with downstream risk. GAR exploits elicitability and strictly consistent scoring rules to cast conditional risk estimation as a regression problem within generative modeling. A minimax formulation with an adversarial policy ensures robustness under policy shift. Empirical results on S\&P 500 data with jointly elicitable (VaR, ES) show that GAR outperforms unconditional generators, DCC--GARCH, and linear direct model baselines in fixed-policy settings, while remaining stable under worst-case policies, in contrast to the degradation observed for fixed-policy trained counterparts.

\clearpage
\printbibliography

\clearpage
\appendix
\section{Configuration}
\label{App-A}
\begin{table}[h]
\centering
\renewcommand{\arraystretch}{1.15}
\begin{tabular}{>{\centering\arraybackslash}m{2.0cm} >{\centering\arraybackslash}m{5.0cm} >{\centering\arraybackslash}m{6.5cm}}
\toprule
& \textbf{Configuration} & \textbf{Values} \\
\midrule

\multirow{4}{*}{Optimization}
& Initial learning rate & $10^{-10}$ \\
& Learning rate schedule & OneCycleLR (\cite{DBLP:journals/corr/abs-1708-07120}) \\
& Batch size & 128 \\
& Optimization algorithm & Adam (\cite{kingma2017adammethodstochasticoptimization}) \\
\midrule

\multirow{3}{*}{Data setup}
& Number of assets & 9 \\
& Conditioning window length & 5 \\
& Generated trajectory length & 10 \\
\midrule

\multirow{4}{2.0cm}[-3.6ex]{\centering Architecture}
& Simple-linear & layers: 2, hidden dim: 4, activation: LeakyReLU \\
& Encoder-linear & layers: 2, hidden dim: 4, activation: LeakyReLU \\
& Encoder-LSTM & LSTM layers: 1, decoder layers: 1, hidden dim: 4 \\
& Adversarial policy & GRU layers: 3, Portfolio cap (total gross exposure, $\kappa$): 1 \\
\midrule

\multirow{2}{*}{Sampling}
& Monte Carlo sample size & 2,000 \\
& Latent noise distribution & $\mathcal{N}(0,\mathrm{I})$ \\
\midrule

\multirow{2}{*}{\makecell[c]{Risk and\\scoring setup}} & $H_1, H_2$ & $H_1(v)=v,\; H_2(e)=s\,\exp(e/s),\; s=2$ \\
& Quantile ($\alpha$) & 0.05 \\

\bottomrule
\end{tabular}
\caption{Model and experiment configuration details.}
\end{table}

\section{Generator and Policy Architectures}
\label{app-B}
This section summarizes the architectural components used in our experiments. We first detail the three generator architectures and then present the policy architecture used for adversarial training. Throughout, a single training sample consists of
a conditioning window $c \in \mathbb{R}^{M \times T_c}$ (past returns for $M$ assets over
$T_c$ days) and a target scenario $y \in \mathbb{R}^{M \times T}$ (future returns over $T$
days). The generator produces $\hat y \in \mathbb{R}^{M \times T}$ from conditioning information $c$ and latent noise $z \in \mathbb{R}^{d_z}$.

\paragraph{Simple-linear}
The Simple-linear generator applies the same feedforward network independently to each asset, acting only along the time dimension. We define per-asset input 
\[
h_{0,j} := [z, c_{j\cdot}] \in \mathbb{R}^{d_z + T_c},
\]
where \(c_{j\cdot}\) denotes the \(j\)-th row of \(c\), i.e., the conditioning sequence for asset \(j\), and $[z, c_{j\cdot}]$ denotes the vector obtained by stacking $z$ and $c_{j\cdot}$. Per-asset output is
\begin{align}
    \hat y_j = W_L \cdot \sigma(W_{L-1} \dots \sigma(W_1 h_{0,j}+b_1)\cdots+b_{L-1})+b_L  \in \mathbb{R}^{T},
\end{align} 
where $\sigma$ is an activation function (e.g., ReLU) and $\gamma := (W_1,\dots,W_L,b_1,\dots,b_L)$
represent all the parameters in the neural network. The full generator output is obtained by stacking rows,
\[
\hat y = G(z,c;\gamma) :=
\begin{bmatrix}
\hat y_1^\top \\
\vdots \\
\hat y_M^\top
\end{bmatrix}
\in \mathbb{R}^{M \times T},
\]
This architecture imposes no
temporal or cross-asset inductive bias beyond the shared per-asset network.

\paragraph{Encoder-linear}
The Encoder-linear generator uses a global encoder over both assets and time, followed by a
decoder. Let $\bar c \in \mathbb{R}^{MT_c}$ denote the vectorization of c. The encoder takes
$h^{\mathrm{E}}_0 := \bar c$ and produces the encoded representation,
\begin{align}
\hat H = \mathrm{E}(\bar c;\phi) := W^{\mathrm{E}}_{L_1} \cdot \sigma (W^{\mathrm{E}}_{L_1-1} \cdots \sigma(W^{\mathrm{E}}_{1} h^{\mathrm{E}}_0+b^E_1)\cdots +b^E_{L_1-1}) + b^{\mathrm{E}}_{L_1}
\in \mathbb{R}^{d_h},
\end{align}
with encoder parameters $\phi := (W^{\mathrm{E}}_1,\dots,W^{\mathrm{E}}_{L_1}, b^{\mathrm{E}}_1,\dots,b^{\mathrm{E}}_{L_1})$. The decoder input is the joint vector $h^{\mathrm{D}}_0 := [\hat H,z] \in \mathbb{R}^{d_h + d_z}$, and the decoder output is
\begin{align}
\hat Y = D(\hat H,z;\theta) := W^{\mathrm{D}}_{L_2} \cdot \sigma (W^{\mathrm{D}}_{L_2-1} \cdots \sigma(W^{\mathrm{D}}_{1} h^{\mathrm{D}}_0+b^D_1)\cdots +b^D_{L_2-1}) + b^{\mathrm{D}}_{L_2} \in \mathbb{R}^{M T},
\end{align}
with parameters $\theta := (W^{\mathrm{D}}_1,\dots,W^{\mathrm{D}}_{L_2}, b^{\mathrm{D}}_1,\dots,b^{\mathrm{D}}_{L_2})$. Then we reshape the output $\hat Y \in \mathbb{R}^{MT}$ to 
$\hat y \in \mathbb{R}^{M \times T}$. This architecture learns a non-linear global encoding of the conditioning panel $c$ while
keeping a simple fully connected decoder.

\paragraph{Encoder-LSTM}
The Encoder–LSTM generator uses a recurrent encoder over time with cross-sectional inputs.
We view the conditioning window as a time series of cross-sectional vectors
\[
c = (c_1,\dots,c_{T_c}), \qquad c_t \in \mathbb{R}^{M},
\]
where $c_t$ collects the $M$ asset returns at time $t$. The encoder is a Long Short-Term
Memory (LSTM) network (\cite{hochreiter1997long}) with input dimension $M$ and hidden
dimension $d_h$ with parameters $\phi$. For each layer $l=1,\dots,L$, the hidden state $h_t^l \in \mathbb{R}^{d_h}$ and cell state $s_t^l \in \mathbb{R}^{d_h}$ are updated as
\begin{align}
h_t^l,s_t^l=f_{\mathrm{LSTM}}(h_{t-1}^l,s_{t-1}^l,x_t^l;\phi)
\end{align}
using the standard LSTM function. $(h_0,s_0)$ default to zeros if not provided. The input $x_t^l$ to layer $l$ is
\[
x_t^l =
\begin{cases}
c_t & \text{for } l = 1, \\
h_t^{l-1} & \text{for } l \geq 2.
\end{cases}
\]

The encoder representation is the final hidden state,
\[
\hat H = \mathrm{E}(c;\phi) := h_{T_c}^L \in \mathbb{R}^{d_h}.
\]
The decoder is identical to the Encoder-linear case: it maps $(\hat H,z)$ to a vector in
$\mathbb{R}^{M T}$ and reshapes it to an $M \times T$ trajectory matrix. This variant allows the encoder to exploit both temporal and cross-asset structure in the
conditioning information.

\paragraph{Adversarial policy}
Recall~\eqref{eq:policy_seq}: for a scenario $y \in \mathbb{R}^{M \times T}$, the policy $\pi_\phi$ outputs action sequences
\begin{equation}
\pi_\phi(y)
=
\big[w_\phi(y_{1:1}),\ w_\phi(y_{1:2}),\ \ldots,\ w_\phi(y_{1:T-1})\big],
\qquad
w_\phi(\cdot) \in \mathbb{R}^M.
\end{equation}
We implement $w_\phi$ using a multi-layer Gated Recurrent Unit (GRU) (\cite{cho2014propertiesneuralmachinetranslation}) with parameters $\phi$. For each layer $l = 1,\ldots,L$ and time $t = 1,\ldots,T-1$, the hidden state $h_t^l \in \mathbb{R}^{d_h}$ satisfies
\begin{equation}
h_t^l = f_{\mathrm{GRU}}(h_{t-1}^l,\, x_t^l;\, \phi),
\end{equation}
with $h_0^l = \mathbf{0}$. We set $d_h = M$ so that the final hidden state can be interpreted as asset-wise raw actions. The input $x_t^l$ to layer $l$ is
\[
x_t^l =
\begin{cases}
y_t & \text{for } l = 1, \\
h_t^{l-1} & \text{for } l \geq 2.
\end{cases}
\]
The raw actions at time $t$ is
\begin{equation}
w_\phi(y_{1:t}) = h_t^L
\quad \text{for } t = 1,\ldots,T-1.
\end{equation}
These raw actions are then normalized before use in the aggregator.

\end{document}